\documentclass[journal]{IEEEtran}
\ifCLASSINFOpdf
\else
\fi
 \usepackage{hyperref}
\usepackage[compatibility=false]{caption}
\usepackage{amsfonts}
\usepackage{amsmath}
\usepackage{url}
\usepackage[numbers]{natbib}
\usepackage{booktabs}
\usepackage{siunitx}
\usepackage{tabularx}
\usepackage{array}
\usepackage{subcaption}
\usepackage{graphicx}
\usepackage{stfloats}   
\usepackage{eso-pic}

\newcommand{\firstpagenotice}{%
  \AddToShipoutPictureBG*{%
    \AtPageLowerLeft{%
      \hspace*{\dimexpr(\paperwidth-\textwidth)/2\relax}%
      \raisebox{1cm}{\parbox{\textwidth}{%
        \centering\footnotesize
        This work has been submitted to the IEEE for possible publication. Copyright may be transferred without notice, after which this version may no longer be accessible.
      }}%
    }%
  }%
}
\begin{document}
%
\title{Efficient Neural-Network-Based High-Resolution Radiative Transfer for CO$_2$ Retrieval, and Application to Interferometric Sensing} 
%
%
%

\author{
Jordan~Lontsi~Tedongmo,
Yann~Ferrec,
Laurence~Croizé,
Pablo~Musé,
Gabriele~Facciolo,
and~Andrés~Almansa%
\thanks{This work has been partially funded by the French-Uruguayan cooperation program under grant number ECOS Sud U25E01 and by IFUMI IRL-CNRS IFUMI-2030. (\textit{Corresponding author: Jordan Lontsi Tedongmo.})}
\thanks{Jordan Lontsi Tedongmo is with DOTA, ONERA -- The French Aerospace Lab, Chemin de la Hunière, 91123 Palaiseau, France, and also with Centre Borelli, CNRS, ENS Paris-Saclay, Université Paris-Saclay, 91190 Gif-sur-Yvette, France (e-mail: jordan.lontsi\_tedongmo@onera.fr).}%
\thanks{Yann Ferrec and Laurence Croizé are with DOTA, ONERA -- The French Aerospace Lab, Chemin de la Hunière, 91123 Palaiseau, France (e-mail: yann.ferrec@onera.fr; laurence.croize@onera.fr).}%
\thanks{Pablo Musé is with the Facultad de Ingeniería (FIng), Universidad de la República (UdelaR), Montevideo, Uruguay, and also with Centre Borelli, CNRS, ENS Paris-Saclay, Université Paris-Saclay, 91190 Gif-sur-Yvette, France (e-mail: pmuse@fing.edu.uy).}%
\thanks{Gabriele Facciolo is with Centre Borelli, CNRS, ENS Paris-Saclay, Université Paris-Saclay, 91190 Gif-sur-Yvette, France, and also with Institut Universitaire de France (e-mail: gabriele.facciolo@ens-paris-saclay.fr).}%
\thanks{Andrés Almansa is with MAP5 UMR 8145, CNRS, Université Paris Cité, F-75006 Paris, France (e-mail: andres.almansa@parisdescartes.fr).}%
}

%
%

\markboth{Journal of \LaTeX\ Class Files,~Vol.~13, No.~9, September~2014}%
{Shell \MakeLowercase{\textit{et al.}}: Bare Demo of IEEEtran.cls for Journals}
%



\maketitle
\firstpagenotice
\markboth{}{}

\begin{abstract}
Studying climate change requires reducing uncertainties in CO\(_2\) and CH\(_4\) emission estimates to better distinguish anthropogenic from natural sources, which motivates spaceborne measurements with improved revisit frequency and spatial coverage. In this context, the Horizon Europe SCARBOn project assesses a low-cost satellite constellation featuring the NanoCarb imaging interferometer as its core sensor for monitoring CO\(_2\) and CH\(_4\) emissions in the atmosphere. However, estimating CO\(_2\) and CH\(_4\) concentrations with high revisit and spatial coverage poses significant challenges: full-physics retrieval algorithms commonly used rely on repeated high-resolution radiative transfer (RT) simulations, which are computationally expensive when using line-by-line RT models. As an alternative, we propose in this study a feedforward multilayer perceptron (MLP) surrogate designed to accurately and efficiently predict top-of-atmosphere radiances in the CO\(_2\) weak band, using a combined mean absolute error (MAE) loss on radiances and RT Jacobians to preserve both spectral accuracy and sensitivity to geophysical parameters. Coupling the MLP-based RT surrogate with the NanoCarb instrumental response yields an efficient and precise forward model for NanoCarb measurements, which shows promising results for CO\(_2\) concentration retrieval.
\end{abstract}


%
\IEEEpeerreviewmaketitle

\section{Introduction}
%
%
%
%

 

Carbon dioxide (CO\(_2\)) and methane (CH\(_4\)) are major anthropogenic greenhouse gases (GHGs) whose emissions require accurate spatio-temporal monitoring,  thereby supporting the mitigation efforts essential to achieving the Paris Agreement’s goal \citep{naser2022evolution}. In practice, while most current and planned satellite missions dedicated to GHGs monitoring achieve high sensitivity to GHG emissions, their spatial resolution or coverage and revisit frequency are still insufficient to detect and quantify anthropogenic emissions with the required precision \citep{gousset2019nanocarb}.

Advancing these goals is one key motivation for the Horizon Europe SCARBOn project, which assesses a low-cost satellite constellation featuring the miniature GHG sensor NanoCarb as its core sensor for monitoring CO\(_2\) and CH\(_4\) emissions in the atmosphere \citep{gousset2019nanocarb}. Estimating GHGs concentrations from NanoCarb measurements then relies on the so-called full-physics retrieval (or inverse) algorithm, in which high-resolution radiative transfer (RT) calculations must be performed over the sensor response domain for each instrument channel to simulate the measurement, according to the sensor's response function. However, commonly used line-by-line radiative transfer models \citep{clough1995line} are highly computationally intensive, posing significant challenges in operational contexts, especially for the NanoCarb concept, where the satellite constellation will generate vastly more data to process.

To overcome the computational burden of physics-based radiative transfer models in retrieval algorithms, several studies have explored deep neural networks to directly emulate sensor measurements~\citep{gao2021efficient,Nanda2019Neural,stegmann2022deep,liu2023physics}. However, few studies target high-resolution radiative transfer in key GHGs absorption bands \citep{Le2020Application}, and its impact on GHGs retrieval accuracy remains largely unexplored. Other works address the inverse problem using neural networks \citep{David2021XCO2,Reuter2025Retrieving}, but such approaches are generally less flexible in response to changes in instrument characteristics or retrieval settings. Efficient surrogate models for high-resolution radiative transfer, therefore, remain highly desirable.

This paper proposes a fast and accurate neural-network surrogate for high-resolution radiative transfer in the CO$_2$ weak absorption band. Integrated into the NanoCarb forward model, it is evaluated for CO$_2$ total-column and surface albedo retrievals, including on realistic airborne simulation scenarios.

\section{Forward and inverse models for CO2 estimation from NanoCarb measurements}
\subsection{NanoCarb instrumental model}\label{sec:Forward_model}

The NanoCarb sensor \citep{gousset2019nanocarb} is a static Fourier-transform imaging spectrometer designed for measuring CO$_2$ and CH$_4$. The instrument combines an array of low-finesse Fabry–Perot (FP) interferometers and a microlens array, producing multiple co-registered views of the same scene, each modulated by a different interferometric state. This architecture enables wide-swath, high-spectral-resolution measurements in snapshot acquisition mode, with a very compact design. The corresponding instrumental model describes the interferometric intensity recorded at each focal-plane pixel as a function of the incoming radiance. This model is derived from the normalized Fabry--Perot transmission, which for monochromatic radiation can be approximated by \citep{gousset2019nanocarb}
\begin{equation}
T_{FP} \approx \frac{1}{1 + M \sin^2 \left( \frac{\varphi}{2} \right)},
\end{equation}
where \(M\) is the finesse term (depending on the
reflectivity at the FP interfaces) and \(\varphi = 2\pi\sigma\delta\) the phase shift for wavenumber \(\sigma\) and optical path difference (OPD) \(\delta\).

The resulting intensity from a single FP plate (at a given OPD) is obtained by integrating the FP-modulated radiance over the spectral band \(\Delta\sigma\), following:
\begin{equation}
I_{\delta} = \eta \int_{\Delta\sigma} \left( \frac{1}{h \mathbf{c} \sigma} \right) T_{\sigma} T_{FP} L_{\sigma} d\sigma, [\text{e/frame/pixel}],
\label{eq:NanoCarb_inst_model}
\end{equation}

with $T_{\sigma}$ the instrument transmission, $\eta$ the optical–radiometric efficiency, $L_{\sigma}$ the incoming spectral radiance for the observed atmospheric column denoted by $c_j$, and $h$, $\mathbf{c}$ Planck's constant and the speed of light in vacuum, respectively. Simulating NanoCarb measurements therefore requires computing high-resolution spectral radiances, which is achieved using a line-by-line radiative transfer model (see Sec.~\ref{sec:simulation_setup}).

For each scene pixel, the measurements acquired through the different Fabry–Perot plates form a partial interferogram $I$, of dimension equal to the number of FP thicknesses ($n_{FP}$), each component corresponding to a distinct OPD
\begin{equation}
\delta(\sigma,i,\theta,T)=2\,n(\sigma,T)\,\varepsilon_i\cos\theta_r,
\end{equation}
depending on both the FP thickness $\varepsilon_i$ and the incidence angle $\theta$, where $\theta_r$ is the refracted angle inside the FP, $T$ the temperature, and $n$ the optical refractive index of the FP material. The selected thicknesses are tuned to the absorption features of the targeted gas—CO$_2$, CH$_4$, or O$_2$, depending on the instrument configuration—yielding interferograms highly sensitive to its concentration, with a reduced sensitivity to the surface and atmospheric parameters.

Following \eqref{eq:NanoCarb_inst_model}, the resulting interferograms correspond to ideal noiseless measurements. To generate realistic NanoCarb observations, noise accounting for photon noise (Poisson distribution) and read-out noise (Gaussian distribution) is added to the simulated interferograms, essential to reproduce the signal degradation encountered in real measurements.
\subsection{Simulation Setup for Realistic NanoCarb Measurements}\label{sec:simulation_setup}

To evaluate the proposed approach under realistic conditions, we simulated NanoCarb airborne measurements from real surface albedo and CO$_2$ concentration maps, under realistic atmospheric conditions. Surface albedo was derived from airborne hyperspectral reflectance data acquired over Toulouse (France), during the AI4GEO/CAMCATT campaign (AISA FENIX sensor) \cite{roupioz2023multi}, while CO$_2$ total-column maps were obtained from the AVIRIS-NG Benchmark Dataset for Carbon Dioxide Plumes \cite{foote2021impact}, which contains airborne observations of plumes from a variety of industrial emission sources. From these data, two distinct scenarios were constructed and resampled to match the expected spatial resolution of NanoCarb. Simulations were performed for an airborne NanoCarb configuration operating at 3\,km altitude with a 15\,m spatial resolution, assuming an aerosol-free AFGL midlatitude-summer atmosphere \cite{anderson1986afgl}. The instrument operates at a high acquisition rate, allowing the same ground pixel $j$ in the scene to be observed from multiple viewing angles during the scene overpass. Each acquisition comprises $64\times64$ spatial pixels, with an 81-sample partial interferogram recorded for each pixel. Assuming a constant one-pixel shift between consecutive acquisitions, each ground pixel is observed $N_v=64$ times from different viewing angles. These measurements are concatenated into a single vector $\mathbf{I}_j=[I_j(1),\ldots,I_j(N_v)]$, where each $I_j(i)$ denotes an 81-point interferogram.
 Radiative transfer simulations were performed with 4AOP \cite{scott1981fast} using HITRAN database \cite{rothman2009hitran}, combined with the NanoCarb instrument and noise models to generate the final NanoCarb measurements.

\subsection{Retrieval algorithm}\label{sec:retrieval_algorithm}

The NanoCarb CO$_2$ instrument is primarily designed to measure the atmospheric CO$_2$ total column. However, because the measured interferograms are also highly sensitive to surface albedo, the latter is jointly retrieved as a secondary product. Other parameters affecting the weak CO$_2$ absorption band (surface pressure, sun–observation geometry, and water vapor and temperature profiles) are assumed to be known. The state vector associated with each ground pixel $j$ of the map to be reconstructed is therefore reduced to the total CO$_2$ column ($X_{CO_2}$) and the surface albedo (A):
$c_j=\left[X_{CO_2}, A\right]^T.$

The retrieval framework follows the Bayesian formulation previously introduced for NanoCarb \citep{dogniaux2022space}, which relies on the classical Optimal Estimation framework \citep{Rodgers2000Inverse}. A unique atmospheric state $\hat{c}_j$ is estimated from the $N_v$ independent views $\mathbf{I}_j$ available for pixel $j$, by maximizing the posterior probability distribution. Denoting by $p(\mathbf{I}_j|c_j)$ the likelihood and by $p(c_j)$ the prior distribution, the Maximum A Posteriori (MAP) estimator is
\begin{equation}
    \hat{c}_j=
\arg\max_{c_j}
\left\{
\log p(\mathbf{I}_j|c_j)
+
\log p(c_j) \right\}.
\end{equation}
The likelihood is evaluated through the NanoCarb forward model
$
I_j(i)=F(c_j;i)+\tilde n,
$
where \(F(\cdot;i)\) denotes the forward model associated with acquisition \(i\), evaluated at the corresponding observation angle, and \(\tilde n\) the corresponding measurement noise, assumed Gaussian with standard deviation \(\sigma(i)\).
Assuming further a Gaussian prior distribution centered on $\bar{c}$ with covariance matrix $S$, the MAP estimator is then equivalent to the minimization of
\begin{equation}
\begin{aligned}
\chi^2(c_j)=&\frac{1}{N_v}
\sum_{i=1}^{N_v}
\frac{1}{\sigma^2(i)}\left\|
I_j(i)-F(c_j;i)
\right\|_2^2
 \\
&+
(c_j-\bar{c})^T S^{-1}(c_j-\bar{c}).
\end{aligned}
\end{equation}
The resulting nonlinear least-squares problem is solved using the Levenberg--Marquardt algorithm, which is well-suited for this class of optimization problems.

\section{Neural Network forward model}
\subsection{Training Data}\label{sec:Training_data}
To train the NN RT model, synthetic data were generated by sampling the main atmospheric and surface parameters governing radiance formation in the CO$_2$ weak band at 3,km altitude under aerosol-free conditions. Inputs include CO$_2$ total column, surface properties (albedo, altitude, pressure), solar–sensor geometry, and vertical profiles of temperature and water vapor. These variables were sampled according to the distributions summarized in Table~\ref{tab:training-params}, with temperature and water vapor atmospheric profiles drawn from the TIGR database, which contains 2,311 representative tropical, mid-latitude, and polar atmospheres~\cite{rothman2009hitran}. RT simulations use full atmospheric profiles, whereas the NN uses a reduced representation based on total-column CO$_2$ scaling and PCA coefficients for temperature and water vapor. Overall, 70,000 scenarios were generated, yielding high-resolution radiance spectra (6175--6250\,cm$^{-1}$, 0.01\,cm$^{-1}$ sampling, 7,499 channels) and their Jacobians with respect to all input parameters.

The NN forward model was evaluated on an independent test set of approximately 10,000 scenarios, generated using 200 TIGR temperature and water-vapor profiles excluded from training, while sampling the remaining parameters as in Table~\ref{tab:training-params} (with albedo uniformly spanning 0.05--0.7). High-resolution radiances and corresponding NanoCarb interferograms were simulated for an end-to-end evaluation.

By combining broad parameter sampling with representative atmospheric profiles from TIGR, the resulting datasets cover a wide range of realistic atmospheric and surface conditions, ensuring good geospatial  representativity.

\begin{figure*}[b!]
    \centering

    \begin{subfigure}[t]{0.305\textwidth}
        \centering
        \includegraphics[width=\linewidth]{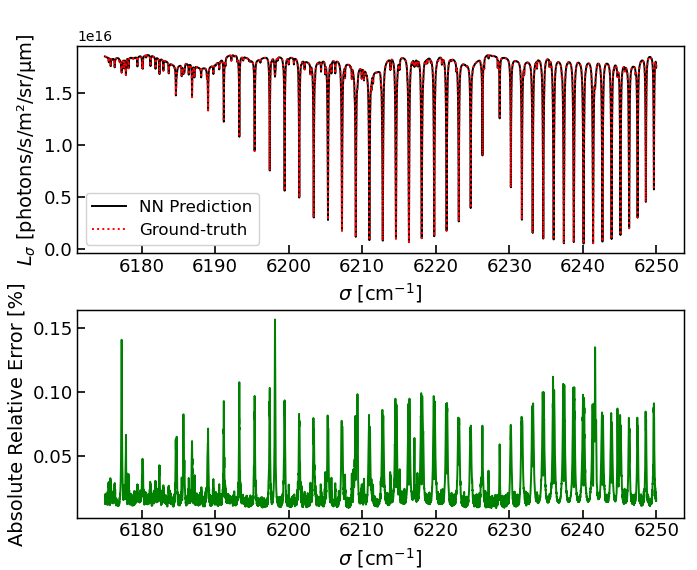}
        \caption*{(a1)}
    \end{subfigure}
    \hfill
    \begin{subfigure}[t]{0.28\textwidth}
        \centering
        \includegraphics[width=\linewidth]{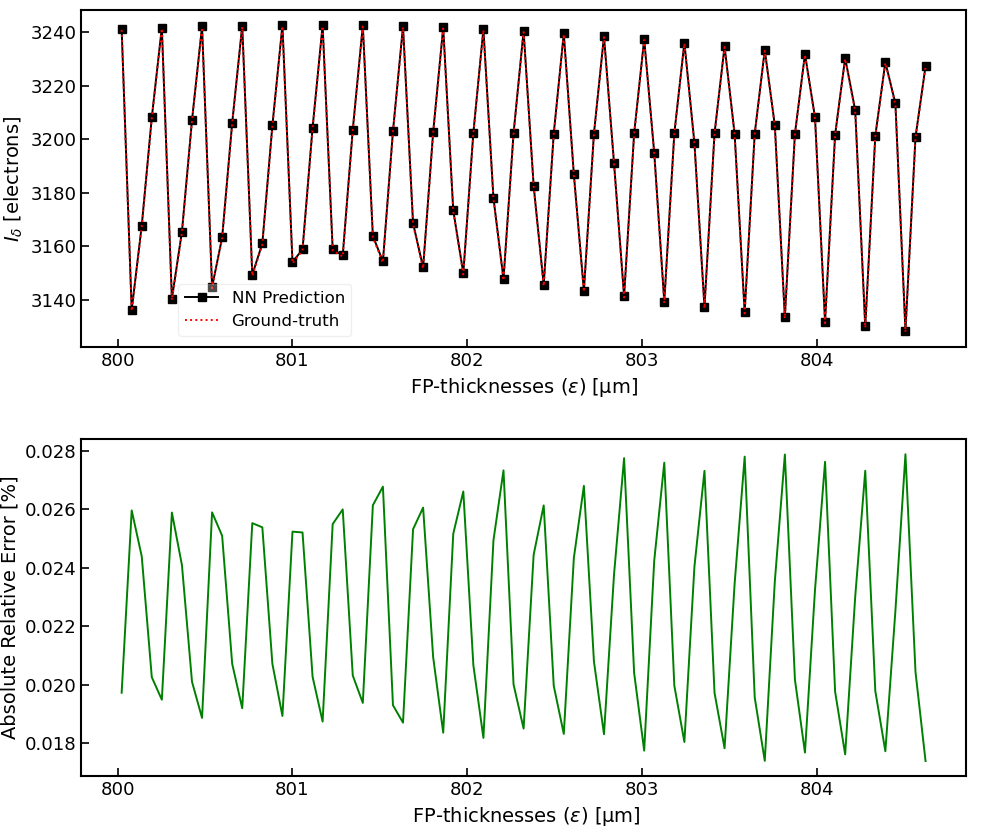}
        \caption*{(a2)}
    \end{subfigure}
    \hfill
    \begin{subfigure}[t]{0.33\textwidth}
        \centering
        \includegraphics[width=\linewidth]{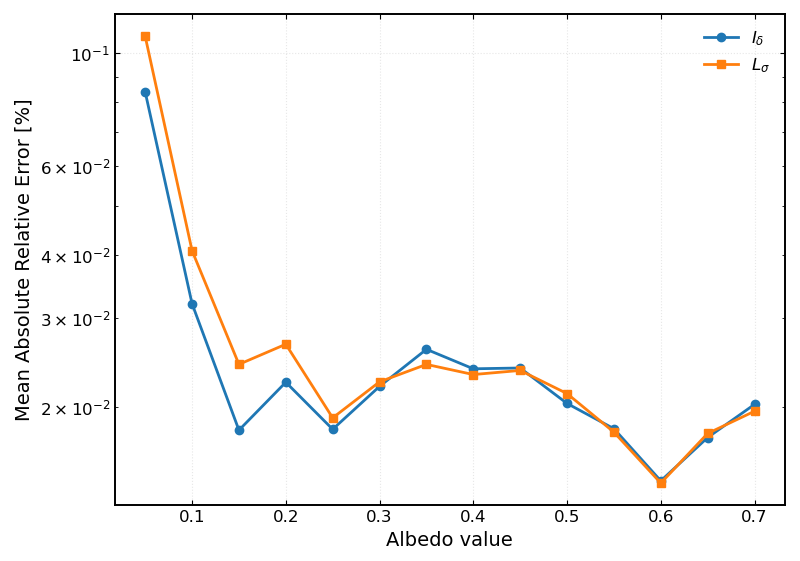}
        \caption*{(b)}
    \end{subfigure}

    \caption{(a1) and (a2): averaged NN predictions vs.\ RT simulations (albedo $= 0.2$) with corresponding averaged absolute relative errors for Radiance and Interferograms. (b): MARE for Radiance (orange) and Interferograms (blue) as a function of albedo.}
    \label{fig:Radiance_Interferogram}
\end{figure*}

\subsection{Neural Network architecture and Training}\label{sec:training}
Our NN radiative transfer surrogate is a feedforward multilayer perceptron (MLP). All inputs are concatenated into a single feature vector $x$ ($n_f = \mathrm{dim}(x)=20$; Table~\ref{tab:training-params}), and the network outputs the full high-resolution radiance spectrum with $n_\sigma=7{,}499$ channels, one per output neuron. Training optimizes the MLP parameters $\theta$ (weights and biases) by minimizing a joint loss on radiances and Jacobians. Inputs and outputs are normalized for stability: $c_j$ denotes the raw input for scenario $j$, with $x_j$ its normalized form; $L_{\sigma}(c_j)$ is the RT-computed radiance spectrum, and $y_F(x_j)$ its normalized counterpart. The NN then predicts a normalized radiance spectrum $y_{\theta}(x_j) \in \mathbb{R}^{n_\sigma}$, with the loss defined as:
\begin{equation}
\small
\mathcal{L}(\theta)
= \frac{1}{N\,n_\sigma}\sum_{j=1}^{N}\sum_{t=1}^{n_\sigma}
\left|y_{\theta}^{(t)}(x_{j})-y_{F}^{(t)}(x_{j})\right|
+ \mathcal{L}_{\mathrm{Jac}}(\theta),
\end{equation}
where $N$ is the number of training samples and $n_\sigma=7{,}499$ the number of spectral channels.

\begin{table}[!t]
\caption{Parameters used to generate the synthetic training dataset for the NN radiative transfer model (without aerosols).}
\label{tab:training-params}
\centering
\footnotesize
\setlength{\tabcolsep}{2pt}
\renewcommand{\arraystretch}{1.0}
\begin{tabularx}{\columnwidth}{l c >{\raggedright\arraybackslash}X c}
\toprule
Parameter & Unit & Distribution & Dim. \\
\midrule
$X_{CO_2}$ & ppm &
$\mathcal{N}(423,\,(0.15\times423)^2)$ &
1 \\

$A$ & -- &
$\log_{10}(A)\sim\mathcal{N}(\log_{10}(0.3),\,(0.7\log_{10}(0.3))^2)$ &
1 \\

$H_{\mathrm{surf}}$ & m &
$\mathcal{N}(300,300^2)$ &
1 \\
$P_{\mathrm{surf}}$ & hPa &
$\mathcal{U}(940,1060)$ &
1 \\
$T(z)$ & K &
2311 TIGR profiles + PCA &
8 \\
$X_{H_2O}(z)$ & g/g &
2311 TIGR profiles + PCA &
6 \\
VZA & deg &
$\mathcal{U}(0,20)$, $1^\circ$ step &
1 \\
SZA & deg &
$\mathcal{U}(0,80)$, $2^\circ$ step &
1 \\
\bottomrule
\end{tabularx}
\end{table}
The use of a Jacobian loss term $\mathcal{L}_{\mathrm{Jac}}$, already introduced in~\cite{liu2023physics}, enforces physically consistent sensitivities to input parameters and also regularizes training:
\begin{equation}
\small
\mathcal{L}_{\mathrm{Jac}}(\theta)
= \sum_{i=1}^{n_f}\alpha_i\,\frac{1}{N\,n_\sigma}\sum_{j=1}^{N}\sum_{t=1}^{n_\sigma}
\left|J_{\theta}^{(t,i)}(x_{j})-J_{F}^{(t,i)}(x_{j})\right|,
\end{equation}
where $J_X^{(t,i)}(x_{j}) = \frac{\partial y_X^{(t)}(x_{j})}{\partial x_j^{(i)}}$ are the Jacobians, computed using finite differences for $J_{F}$ and forward-mode automatic differentiation~\cite{pytorch2023forwardad} for $J_{\theta}$.  The weights $\alpha_i$ balance the contribution of each input parameter $x_i$, {defined from the mean radiance sensitivities to input parameters, as $\alpha_i \propto {y_{\text{mean}}}/{\lvert J_{\text{mean}}^{(i)} \rvert}$ (with $y_{\text{mean}}$ the mean radiance and $J_{\text{mean}}^{(i)}$ the mean Jacobian magnitude w.r.t. $x_i$).}

For the MLP architecture, the best trade-off between accuracy and computational efficiency was obtained with three hidden layers of 100\,/\,500\,/\,3,500 neurons and ELU activations.

\section{Results}
\begin{figure}[t]
    \centering
    \begin{subfigure}[t]{0.48\columnwidth}
        \centering
        \includegraphics[width=\linewidth]{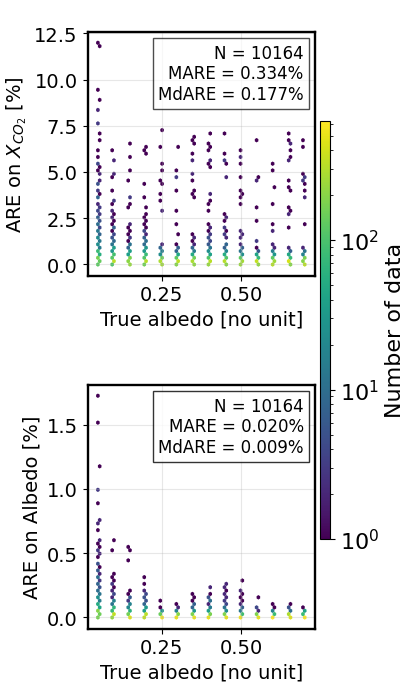}
        \caption{}
        \label{fig:co2_retrievals_hexbin}
    \end{subfigure}
    \hfill
    \begin{subfigure}[t]{0.48\columnwidth}
        \centering
        \includegraphics[width=\linewidth]{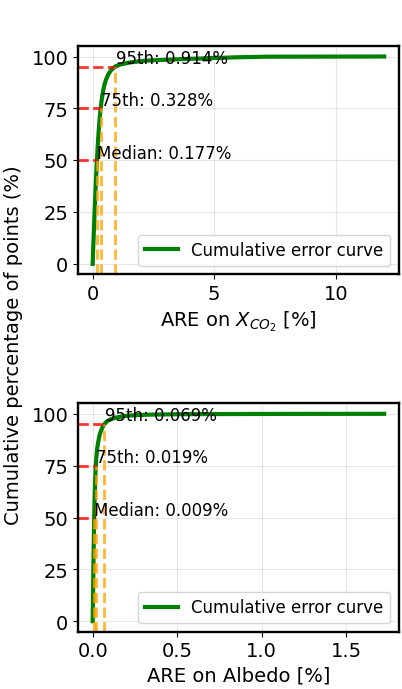}
        \caption{}
        \label{fig:co2_retrievals_cumulative}
    \end{subfigure}
    \caption{Density histograms (a) and cumulative distributions (b) of CO2$_2$
total column and albedo retrieval errors, estimated over $\approx10{,}000$ independent test cases.}
    \label{fig:co2_retrievals}
\end{figure}

\begin{figure}[t]
    \centering
    \setlength{\tabcolsep}{2pt}
    \begin{tabular}{cc}
        \textbf{~~~~4A/OP } & \textbf{NN } \\[2pt]
        \includegraphics[width=0.45\linewidth]{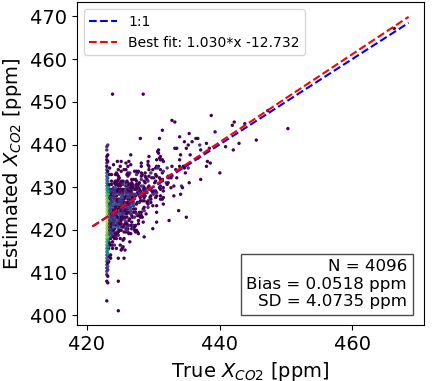} &
        \includegraphics[width=0.49\linewidth]{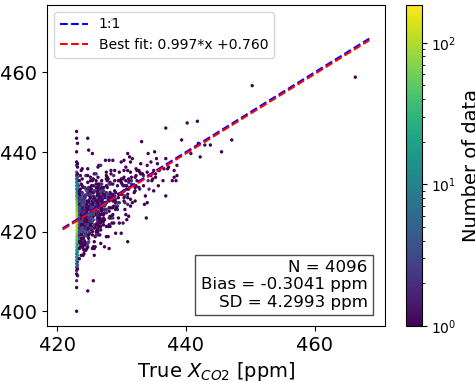} \\[4pt]
        \includegraphics[width=0.45\linewidth]{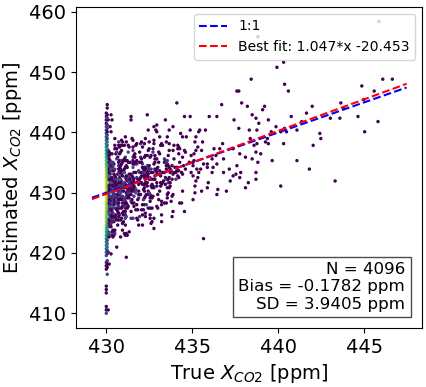} &
        \includegraphics[width=0.5\linewidth]{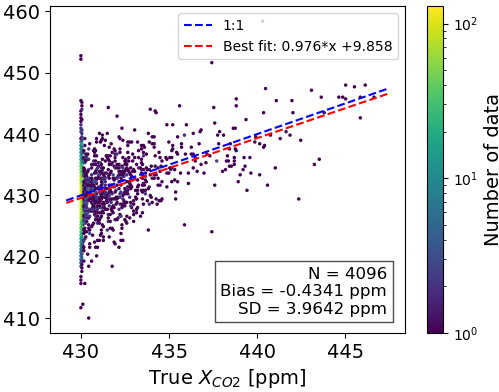} \\
    \end{tabular}
    \caption{
        Scatter plots of estimated vs.\ reference $X_{\text{CO}_2}$ for scenario 1 (top) and scenario 2 (bottom), with 4A/OP retrieval (left) and NN retrieval (right); corresponding maps in Fig.~\ref{fig:gt_vs_retrievals}.
    }
    \label{fig:scatter_co2}
\end{figure}

\begin{figure}[t]
    \centering
    \setlength{\tabcolsep}{1pt}
    \begin{tabular}{ccc}
        \textbf{~~~~Reference} & \textbf{4A/OP Retrieval} & \textbf{NN Retrieval} \\[2pt]
        \includegraphics[width=0.38\linewidth]{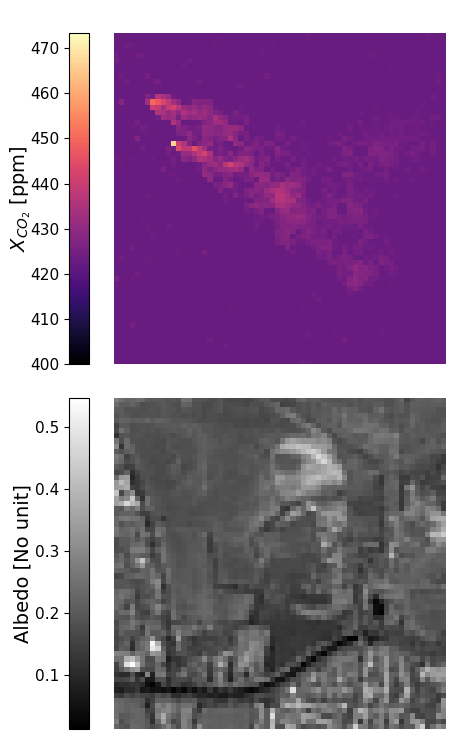} &
        \includegraphics[width=0.285\linewidth]{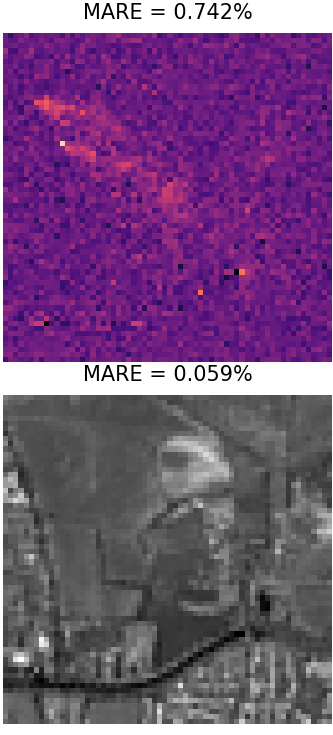} &
        \includegraphics[width=0.294\linewidth]{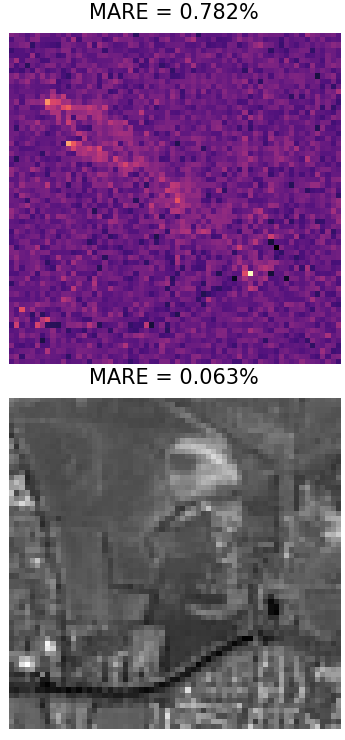} \\[4pt]
        \includegraphics[width=0.38\linewidth]{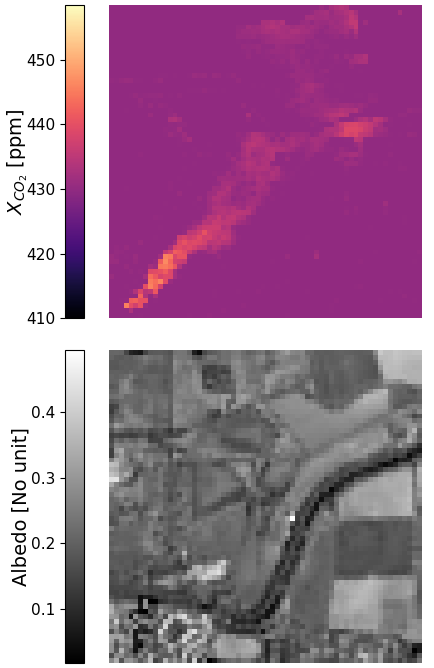} &
        \includegraphics[width=0.285\linewidth]{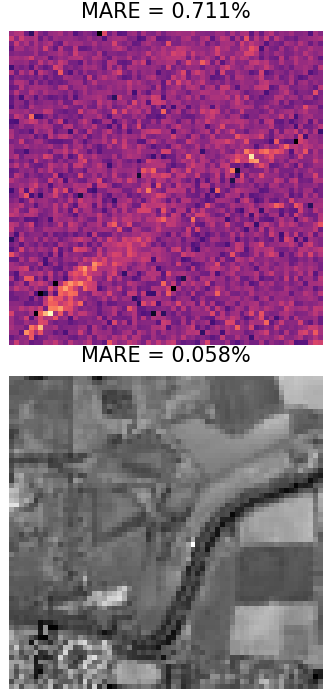} &
        \includegraphics[width=0.288\linewidth]{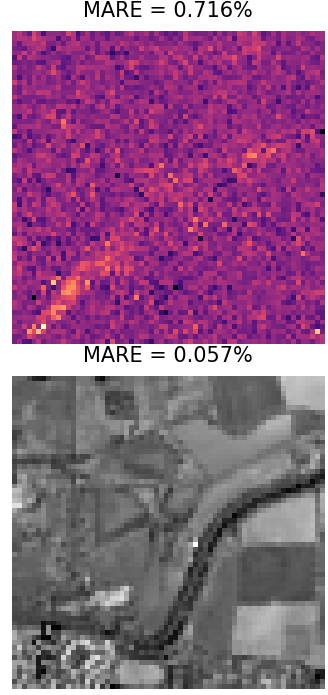} \\
    \end{tabular}
    \caption{
        Estimated $X_{\text{CO}_2}$ and albedo maps for scenario 1 (rows 1–2) and scenario 2 (rows 3–4). \textit{Left:} Reference (simulated true state). \textit{Center:} 4A/OP retrieval. \textit{Right:} NN retrieval. MARE values are indicated above each estimated map.}
    \label{fig:gt_vs_retrievals}
\end{figure}

We first evaluate the accuracy of the trained NN RT surrogate together with the resulting NN-based NanoCarb forward model, obtained by combining the NanoCarb instrumental model with the NN RT model as in \eqref{eq:NanoCarb_inst_model}. Using~$\approx10{,}000$ independently simulated test cases, the proposed model accurately reproduces both radiances and partial interferograms. As summarized in Fig.~\ref{fig:Radiance_Interferogram} (b), the overall mean absolute relative error (MARE) remains below 0.04\% across all channels, reaching at most 0.1\% for the lowest-albedo cases.

We then quantify the impact of this approximation on the retrieval accuracy by replacing the full-physics RT model with the NN surrogate within the inversion framework (Sec.~\ref{sec:retrieval_algorithm}) and retrieving $X_{CO_2}$ and surface albedo from the same $\approx10{,}000$ noise-free simulated NanoCarb measurements. As shown in Fig.~\ref{fig:co2_retrievals}, the surrogate introduces negligible retrieval errors: mean (median) relative errors of 0.33\% (0.17\%) for $X_{CO_2}$ and 0.02\% (0.009\%) for surface albedo. These results are obtained while reducing the forward-model computation time by up to 150× on CPU (Intel Xeon E5-2650 v4 @ 2.20,GHz) and more than 500× on GPU (NVIDIA Quadro RTX 6000).

Finally, we validate the complete retrieval framework on the realistic airborne scenarios described in Sec.~\ref{sec:simulation_setup}. The retrieved albedo maps closely match the references, with only marginal differences between the full-physics and NN-based forward models. For $CO_2$, the full-physics retrieval achieves the expected performance under the assumed instrument noise level (equivalent to 4\,ppm), while the NN-based retrieval reproduces the same concentration patterns without noticeable artifacts (Fig.~\ref{fig:gt_vs_retrievals}). As confirmed by the scatter plots (Fig.~\ref{fig:scatter_co2}), the additional bias and standard deviation from the NN approximation remain below 0.5\,ppm for both scenarios, consistent with the 0.17\% median retrieval error over the $\approx10{,}000$ test cases. This accuracy comes with a drastic reduction in computation time, from nearly one month with the full-physics model to only a few hours on GPU.

\section{Conclusions and Future Work}
This work demonstrates the potential of neural-network surrogates for radiative transfer modeling in greenhouse-gas absorption bands, providing a flexible framework to substantially accelerate the processing of measurements. Despite large computational gains, approximation errors remain low, with radiance errors below 0.04\% on average and retrieval mean errors of only 0.33\% for CO$_2$ and 0.02\% for surface albedo in our NanoCarb test case. Validation on realistic airborne scenarios confirms the approach's suitability under operational conditions, reproducing full-physics CO$_2$ estimates with an additional bias and standard deviation well below 0.5\,ppm, while reducing processing time from nearly a month to a few hours on GPU. These results highlight the strong potential of NN surrogates for operational atmospheric retrievals, and the proposed methodology can readily be extended to other SWIR greenhouse-gas bands and sensors. Future work will focus on accounting for aerosol effects and incorporating radiative-transfer physics directly into network training, an increasingly explored direction for improving accuracy and physical consistency. More broadly, this combination of computational efficiency, compact models, and negligible approximation errors paves the way for more advanced retrieval strategies.


\bibliography{bibtex/bib/IEEEexample}

\begin{thebibliography}{10}
\providecommand{\url}[1]{#1}
\csname url@samestyle\endcsname
\providecommand{\newblock}{\relax}
\providecommand{\bibinfo}[2]{#2}
\providecommand{\BIBentrySTDinterwordspacing}{\spaceskip=0pt\relax}
\providecommand{\BIBentryALTinterwordstretchfactor}{4}
\providecommand{\BIBentryALTinterwordspacing}{\spaceskip=\fontdimen2\font plus
\BIBentryALTinterwordstretchfactor\fontdimen3\font minus
  \fontdimen4\font\relax}
\providecommand{\BIBforeignlanguage}[2]{{%
\expandafter\ifx\csname l@#1\endcsname\relax
\typeout{** WARNING: IEEEtran.bst: No hyphenation pattern has been}%
\typeout{** loaded for the language `#1'. Using the pattern for}%
\typeout{** the default language instead.}%
\else
\language=\csname l@#1\endcsname
\fi
#2}}
\providecommand{\BIBdecl}{\relax}
\BIBdecl

\bibitem{naser2022evolution}
M.~M. Naser and P.~Pearce, ``Evolution of the international climate change
  policy and processes: Unfccc to paris agreement,'' in \emph{Oxford Research
  Encyclopedia of Environmental Science}, 2022.

\bibitem{gousset2019nanocarb}
S.~Gousset, L.~Croiz{\'e}, E.~Le~Coarer, Y.~Ferrec, J.~Rodrigo-Rodrigo,
  L.~Brooker, and S.~consortium http://scarbo-h2020. eu/, ``Nanocarb
  hyperspectral sensor: on performance optimization and analysis for greenhouse
  gas monitoring from a constellation of small satellites,'' \emph{CEAS Space
  Journal}, vol.~11, no.~4, pp. 507--524, 2019.

\bibitem{clough1995line}
S.~A. Clough and M.~J. Iacono, ``Line-by-line calculation of atmospheric fluxes
  and cooling rates: 2. application to carbon dioxide, ozone, methane, nitrous
  oxide and the halocarbons,'' \emph{Journal of Geophysical Research:
  Atmospheres}, vol. 100, no.~D8, pp. 16\,519--16\,535, 1995.

\bibitem{gao2021efficient}
M.~Gao, B.~A. Franz, K.~Knobelspiesse, P.-W. Zhai, V.~Martins, S.~Burton,
  B.~Cairns, R.~Ferrare, J.~Gales, O.~Hasekamp \emph{et~al.}, ``Efficient
  multi-angle polarimetric inversion of aerosols and ocean color powered by a
  deep neural network forward model,'' \emph{Atmospheric Measurement Techniques
  Discussions}, vol. 2021, pp. 1--41, 2021.

\bibitem{Nanda2019Neural}
S.~Nanda, M.~de~Graaf, M.~Sneep, J.~F. de~Haan, P.~Stammes, A.~F.~J. Sanders,
  O.~N.~E. Tuinder, J.~P. Veefkind, and P.~F. Levelt, ``A neural network
  radiative transfer model approach applied to the {Tropospheric Monitoring
  Instrument} aerosol height algorithm,'' \emph{Atmospheric Measurement
  Techniques}, vol.~12, no.~12, pp. 6619--6634, 2019.

\bibitem{stegmann2022deep}
P.~G. Stegmann, B.~Johnson, I.~Moradi, B.~Karpowicz, and W.~McCarty, ``A deep
  learning approach to fast radiative transfer,'' \emph{Journal of Quantitative
  Spectroscopy and Radiative Transfer}, vol. 280, p. 108088, 2022.

\bibitem{liu2023physics}
Q.~Liu and X.~Liang, ``Physics constraint deep learning based radiative
  transfer model,'' \emph{Optics Express}, vol.~31, no.~17, pp.
  28\,596--28\,610, 2023.

\bibitem{Le2020Application}
T.~Le, C.~Liu, B.~Yao, V.~Natraj, and Y.~L. Yung, ``Application of machine
  learning to hyperspectral radiative transfer simulations,'' \emph{Journal of
  Quantitative Spectroscopy and Radiative Transfer}, vol. 246, p. 106928, 2020.

\bibitem{David2021XCO2}
L.~David, F.-M. Br{\'e}on, and F.~Chevallier, ``{XCO$_2$} estimates from the
  {OCO-2} measurements using a neural network approach,'' \emph{Atmospheric
  Measurement Techniques}, vol.~14, no.~1, pp. 117--132, 2021.

\bibitem{Reuter2025Retrieving}
M.~Reuter, M.~Buchwitz, O.~Schneising, S.~Noel, H.~Bovensmann, and J.~P.
  Burrows, ``Retrieving the atmospheric concentrations of carbon dioxide and
  methane from the {European Copernicus CO2M} satellite mission using
  artificial neural networks,'' \emph{Atmospheric Measurement Techniques},
  vol.~18, no.~1, pp. 241--264, 2025.

\bibitem{roupioz2023multi}
L.~Roupioz, X.~Briottet, K.~Adeline, A.~Al~Bitar, D.~Barbon-Dubosc,
  R.~Barda-Chatain, P.~Barillot, S.~Bridier, E.~Carroll, C.~Cassante
  \emph{et~al.}, ``Multi-source datasets acquired over toulouse (france) in
  2021 for urban microclimate studies during the camcatt/ai4geo field
  campaign,'' \emph{Data in Brief}, vol.~48, p. 109109, 2023.

\bibitem{foote2021impact}
M.~D. Foote, P.~E. Dennison, P.~R. Sullivan, K.~B. O'Neill, A.~K. Thorpe, D.~R.
  Thompson, D.~H. Cusworth, R.~Duren, and S.~C. Joshi, ``Impact of
  scene-specific enhancement spectra on matched filter greenhouse gas
  retrievals from imaging spectroscopy,'' \emph{Remote Sensing of Environment},
  vol. 264, p. 112574, 2021.

\bibitem{anderson1986afgl}
G.~P. Anderson, S.~A. Clough, F.~Kneizys, J.~H. Chetwynd, and E.~P. Shettle,
  ``Afgl atmospheric constituent profiles (0.120 km),'' Tech. Rep., 1986.

\bibitem{scott1981fast}
N.~Scott and A.~Chedin, ``A fast line-by-line method for atmospheric absorption
  computations: The automatized atmospheric absorption atlas,'' \emph{Journal
  of Applied Meteorology (1962-1982)}, pp. 802--812, 1981.

\bibitem{rothman2009hitran}
L.~S. Rothman, I.~E. Gordon, A.~Barbe, D.~C. Benner, P.~F. Bernath, M.~Birk,
  V.~Boudon, L.~R. Brown, A.~Campargue, J.-P. Champion \emph{et~al.}, ``The
  hitran 2008 molecular spectroscopic database,'' \emph{Journal of Quantitative
  Spectroscopy and Radiative Transfer}, vol. 110, no. 9-10, pp. 533--572, 2009.

\bibitem{dogniaux2022space}
M.~Dogniaux, C.~Crevoisier, S.~Gousset, {\'E}.~Le~Coarer, Y.~Ferrec,
  L.~Croiz{\'e}, L.~Wu, O.~Hasekamp, B.~Sic, and L.~Brooker, ``The space carbon
  observatory (scarbo) concept: assessment of x co 2 and x ch 4 retrieval
  performance,'' \emph{Atmospheric Measurement Techniques}, vol.~15, no.~16,
  pp. 4835--4858, 2022.

\bibitem{Rodgers2000Inverse}
C.~D. Rodgers, \emph{Inverse Methods for Atmospheric Sounding: Theory and
  Practice}.\hskip 1em plus 0.5em minus 0.4em\relax World Scientific, 2000,
  vol.~2.

\bibitem{pytorch2023forwardad}
\BIBentryALTinterwordspacing
{PyTorch Team}. (2023, Updated: Apr 18) Forward-mode automatic differentiation
  (beta). PyTorch Documentation. Accessed: 2025-09-05. [Online]. Available:
  \url{https://pytorch.org/docs/stable/tutorials/intermediate/forward_ad_usage.html}
\BIBentrySTDinterwordspacing

\end{thebibliography}
\bibliographystyle{IEEEtran}

\newpage


\ifCLASSOPTIONcaptionsoff
  \newpage
\fi

\end{document}